\pdfoutput=1

\documentclass[11pt]{article}

\usepackage[final]{acl}

\usepackage{times}
\usepackage{latexsym}

\usepackage[T1]{fontenc}

\usepackage[utf8]{inputenc}

\usepackage{microtype}

\usepackage{inconsolata}

\usepackage{graphicx}

\usepackage{times}
\usepackage{soul}
\usepackage{multirow} 
\usepackage{url}
\usepackage{tabularx} 
\usepackage{algorithmic}
\usepackage{amssymb}
\usepackage{amsmath}
\usepackage[switch]{lineno}
\usepackage{stfloats}
\usepackage{makecell}
\usepackage{tcolorbox}

\usepackage[ruled,linesnumbered]{algorithm2e}
\SetKwInput{KwInput}{Input}                
\SetKwInput{KwOutput}{Output}              

\usepackage{longtable}            
\usepackage{listings}
\usepackage{supertabular,booktabs}

\title{Towards Generating Controllable and Solvable Geometry Problem by Leveraging Symbolic Deduction Engine}

\author{
 \textbf{Zhuoxuan Jiang\textsuperscript{1}},
 \textbf{Tianyang Zhang\textsuperscript{2}},
 \textbf{Peiyan Peng\textsuperscript{2}},
 \textbf{Jing Chen\textsuperscript{3}},
\\
 \textbf{Yinong Xun\textsuperscript{2}},
 \textbf{Haotian Zhang\textsuperscript{2}},
 \textbf{Lichi Li\textsuperscript{4}},
 \textbf{Yong Li\textsuperscript{5}},
 \textbf{Shaohua Zhang\textsuperscript{1}}
\\
 \textsuperscript{1}Shanghai Business School, Shanghai, China \\
 \textsuperscript{2}Learnable.ai, Shanghai, China \\
 \textsuperscript{3}Shanghai Jiaotong University, Shanghai, China \\
 \textsuperscript{4}Cisco Systems Inc., San Francisco, CA, USA \\
 \textsuperscript{5}Beijing Shangruitong Education Technology Co., Ltd. (TeacherClub.com), Beijing, China \\
 \small{
   \texttt{jzx@sbs.edu.cn}, \texttt{tzhang@aggies.ncat.edu}
 }
}

\begin{document}

\maketitle

\begin{abstract}
Generating high-quality geometry problems is both an important and challenging task in education. Compared to math word problems, geometry problems further emphasize multi-modal formats and the translation between informal and formal languages. In this paper, we introduce a novel task for geometry problem generation and propose a new pipeline method: the Symbolic Deduction Engine-based Geometry Problem Generation framework (SDE-GPG). The framework leverages a symbolic deduction engine and contains four main steps: (1) searching a predefined mapping table from knowledge points to extended definitions, (2) sampling extended definitions and performing symbolic deduction, (3) filtering out unqualified problems, and (4) generating textual problems and diagrams.
Specifically, our method supports to avoid inherent biases in translating natural language into formal language by designing the mapping table, and guarantees to control the generated problems in terms of knowledge points and difficulties by an elaborate checking function. With obtained formal problems, they are translated to natural language and the accompanying diagrams are automatically drew by rule-based methods.
We conduct experiments using real-world combinations of knowledge points from two public datasets. The results demonstrate that the SDE-GPG can effectively generate readable, solvable and controllable geometry problems. 
\end{abstract}

\section{Introduction}


In the field of education, developing an automatic problem generation tool is valuable for both teachers and students. Teachers or problem designers can use the tool to save time and effort, enhancing the efficiency of the problem production process~\cite{wang2021math,cao2022disk}. Meanwhile, students can leverage the tool to generate personalized problems based on their background and interests, improving their learning outcomes~\cite{polozov2015personalized,bernacki2018role}. 
In this paper, the research objective is to investigate how to generate geometry problems which are always less-studied before, to our best knowledge.

Current related studies primarily focus on the generation of math word problems~\cite{qin2023mathematical,christ2024mathwell,liu2024comet,qin2024math}. 
Intuitively, different types of mathematical problems are designed to assess various educational abilities. For example, math word problems emphasize language understanding, mathematical modeling, and equation deduction, while geometry problems require spatial imagination, calculation and reasoning skills, as well as mastery of geometric theorems and properties~\cite{liu2020mathematical}. Therefore, although both types of problems prioritize readability in natural language and solvability, methods for generating math word problems cannot be directly applied to geometry problems. Specifically, based on our observation, generating a geometry problem necessitates supporting a strict, step-by-step reasoning process based on geometric theorems, often in formal language, and requires multi-modal capabilities to present the problem in both textual and visual forms. These factors make geometry problem generation more challenging.

\begin{figure}
\centerline{\includegraphics[width=0.5\textwidth]{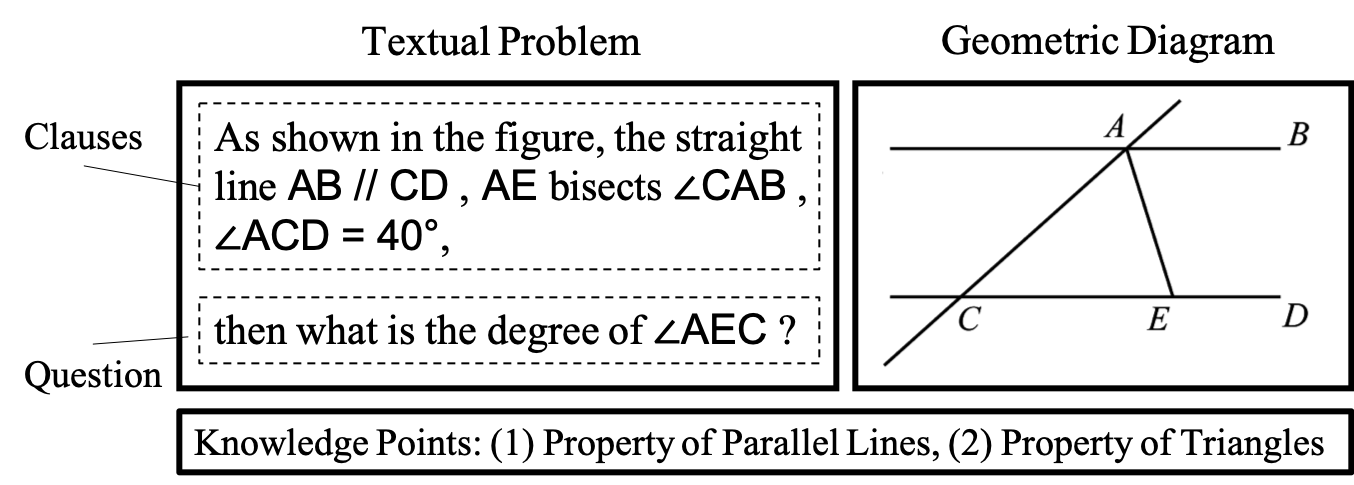}}
\caption{A typical geometry problem consists of a paragraph of textual problem and a geometric diagram. The textual problem is made up of clauses and a question, combining mathematical expressions with natural language. The diagram is sometimes not required.} \label{sample}
\end{figure}

To be more specific, as shown in Figure~\ref{sample}, a typical geometry problem consists of a paragraph of \textit{textual problem} and an accompanying \textit{geometric diagram}. Within the paragraph of textual problem, the text is a mixture of mathematical expressions (e.g., [$AB \parallel CD$]) and natural language (e.g., [As shown in the figure...]). Aside from the final \textit{question} sentence (e.g., [then what is the degree of $\angle AEC$?]), all other textual content are \textit{clauses}. To solve the problem, appropriate geometric \textit{knowledge points}\footnote{Geometric knowledge points, also referred to as geometric rules, include theorems and properties. We do not distinguish between them in the remainder of this paper.} (e.g., the properties of parallel lines and triangles in the case of Figure~\ref{sample}) should be applied during the reasoning process from clauses to the question. If there exists at least one such strict and step-by-step reasoning path, we believe that the geometry problem can be called solvable.

Following the existing studies on controllable problem generation~\cite{liu2024comet}, we also consider several analogous control variables as input, such as the knowledge points and difficulty degree. In summary, to generate controllable high-quality geometry problems, several basic elements should be involved during method design: (1) the textual problem, including clauses and a question, (2) a geometric diagram, and (3) an answer presented as a step-by-step reasoning path. Most importantly, the generated problems must be rightly solvable. Thus, the proposed task definition is that to generate a geometry problem, the knowledge points and difficulty as control variables are given, and the above-mentioned three basic elements would be outputted. In this paper, considering the complexity of the whole geometric domain, we focus on Euclidean plane geometry, leaving the exploration of topics such as geometric inequalities and combinatorial geometry for future work. The following Section \ref{pd} (Problem Definition) will introduce a detailed description of the proposed task.

To achieve the task of geometry problem generation, with a focus on readability, solvability, and controllability, we propose a pipeline method called the Symbolic Deduction Engine-based Geometry Problem Generation framework (SDE-GPG). The framework consists of four main steps: (1) searching a knowledge point-to-extended definition mapping table, (2) sampling extended definitions and performing symbolic deduction, (3) filtering out unqualified problems, and (4) generating textual problems and geometric diagrams. The details of SDE-GPG is introduced in the Section \ref{method} (Method).
In order to evaluate the effectiveness of our proposed method, we manually curate two public datasets containing real-world combinations of knowledge points. This approach helps avoid invalid combinations, as using arbitrary knowledge points sometimes results in unsolvable conclusion. After thorough human evaluation, we find that the generated problems by our method ensure decent solvability and good consistency with control variables, along with precise descriptions in both natural language and visual diagrams. Due to the limited space, the part of related work is put into the Section \ref{appendix} (Appendix).

The contributions of this paper include:

\begin{itemize}
\setlength{\itemsep}{0pt}
  \setlength{\parskip}{0pt}
  \setlength{\itemindent}{0em}
    \item We propose a \textbf{new, simplified task definition} for generating geometry problems. Controlled by \textbf{knowledge points and difficulty degree}, this task outputs \textbf{readable and solvable problems}. Each problem consists of three components: (1) a paragraph of textual clauses and question, (2) a geometric diagram, and (3) a step-by-step reasoning path as the answer.
    \item We leverage a symbolic deduction engine and propose a pipeline framework to accomplish the task, called the \textbf{Symbolic Deduction Engine-based Geometry Problem Generation framework (SDE-GPG)}. The framework consists of four steps: (1) searching a knowledge point-to-exDefinition mapping table, (2) sampling exDefinitions and performing symbolic deduction, (3) filtering out unqualified problems, and (4) generating textual problems and diagrams.
    \item We collect \textbf{two datasets} and conduct thorough experiments to evaluate the \textbf{readability, solvability} and \textbf{controllability} of the generated problems. The experimental results demonstrate the effectiveness of our method in terms of all the aspects. The code, data, templates and other resources are public to facilitate the successive researches\footnote{\url{https://github.com/tianyangzhang123/SDE-GPG-ACL25}}. 
\end{itemize}

\section{Related Work}

\subsection{Educational Question Generation}

Educational problem generation is a broad topic, as different subjects and problem types may focus on specific pedagogical objectives~\cite{gorgun2024instruction}. In the field of mathematics, current studies primarily focus on generating math word problems, with two main research lines: controllable generation and analogy generation~\cite{liu2024comet}. In controllable generation, problems are created based on parameters such as knowledge points~\cite{wu2022automatic}, grade~\cite{qin2024math}, difficulty level~\cite{jiao2023automatic,hwang2024using}, and more~\cite{wang2021math,cao2022disk}. In analogy generation, problems are generated by starting with a seed problem~\cite{zhou2023learning,norberg2023rewriting}. Additionally, some research has focused on generating multi-modal math word problems~\cite{liu2024comet}. Recently, the educational value of generated math problems has gained significant attention, with studies examining factors like `age-appropriateness'~\cite{christ2024mathwell} and `cone of experience'~\cite{liu2024comet}. However, despite these advancements, to the best of our knowledge, the generation of geometry problems remains unexplored. This paper presents a pioneering study on generating such problems.

\subsection{Geometric Synthetic Data Augmentation}

Our task is related to the field of geometry synthetic data augmentation, which is a promising direction for generating large amounts of high-quality data to train theorem provers and verifiers~\cite{firoiu2021training,wang2023lego,azerbayev2023llemma,yang2024leandojo}. Early studies primarily focused on generating synthetic proofs for existing, human-curated problems~\cite{polu2022formal,lample2022hypertree}. Recently, AlphaGeometry has made a notable contribution on end-to-end generating vast amounts of geometric reasoning data by using a symbolic deduction engine (SDE) and uses the data to train an LLM for problem solving~\cite{trinh2024solving}. Inspired by AlphaGeometry, we leverage the SDE framework to generate solvable geometry problems. The largest difference between these works and ours is that they are for data augmentation to train LLMs, while we should focus more on the problem quality and controllability for the purpose of educational significance.

\subsection{Formal Language for Geometry}

In the field of mathematics, various formal languages have been proposed for automated geometric theorem proving, such as Lean~\cite{de2015lean,moura2021lean}, and several provers and reasoners have been developed using the languages like JGEX~\cite{ida2013automated}, GEX~\cite{chou2000deductive} and LeanReasoner~\cite{raffel2020exploring}. When using formal languages, theorems and proofs are typically encoded in a machine-verifiable format, and rigorous logical rules are applied to ensure the correctness of reasoning. However, fully automated provers still face challenges in autoformalization, which refers to the automatic conversion of informal language into machine-readable formal statements. Early approaches use neural machine translation to map LaTeX-formatted texts to formal languages~\cite{wang2018first,bansal2020learning,cunningham2023towards}. Recently, LLMs and in-context learning~\cite{brown2020language} have expanded the possibilities in this area~\cite{wu2022autoformalization,agrawal2022towards,gadgil2022towards,murphy2024autoformalizing}. Beyond translation-based methods, some structured frameworks have been introduced~\cite{patel2023new,ying2024lean,poiroux2024improving}, while DSP~\cite{jiang2022draft} and its variant~\cite{zhao2024subgoalxl} leverage Minerva~\cite{lewkowycz2022solving} to generate informal proofs that are later converted into formal proof sketches. Despite these advancements, autoformalization still struggles to achieve fully correct translation from natural language to formal language. It is notable that the translation from formal language to natural language and diagrams is generally error-tolerant and deterministic~\cite{trinh2024solving}, and we leverage the characteristics for our task.

\section{Problem Definition}\label{pd}

In this section, we present the problem definition. The terms and notations can be referred to Table~\ref{table:term} of the Appendix.

\paragraph{\textsc{Definition 1}:
Knowledge Point and Difficulty Degree.} The geometric \textit{knowledge points} refer to geometric theorems and properties, denoted as $\mathcal{K}=\{K_1, K_2, \dots, K_{N_k}\}$. For example, $K_1$, which is [$\text{perp}\,a\,b\,c\,d, \text{perp}\,c\,d\,e\,f, \text{ncoll}\,a\,b\,e \Rightarrow \text{para}\,a\,b\,e\,f$], means the parallel line determination theorem. The \textit{difficulty degree} is set as three levels, i.e., Easy, Moderate and Difficult, in this paper.

\paragraph{\textsc{Definition 2}: Premise, Conclusion and Definition.} Each knowledge point $K_i$ consists of a set of \textit{premises} $P_i$ and a \textit{conclusion} $C_i$, denoted as $K_i = \{P_i, C_i\}$. For example, for $K_1$, we have $P_1 = \{\text{perp}\,a\,b\,c\,d, \text{perp}\,c\,d\,e\,f, \text{ncoll}\,a\,b\,e\}$ and $C_1 = \{\text{para}\,a\,b\,e\,f\}$. To start a symbolic deduction engine, the \textit{definitions}, denoted as $\mathcal{D} = \{D_1, D_2, \dots, D_{N_d}\}$, are essential to provide a complete description of a geometry, while the $\mathcal{K}$ are selectively used for reasoning. The premises, conclusions, and definitions are all expressed in formal language.



\paragraph{\textsc{Definition 3}: Knowledge Point-to-exDefinition Mapping Table (K2exD-MT).}  
We define the combination of any definitions as \textit{extended definitions} (exDefinition), denoted as $ex\mathcal{D}=\{f_{\text{minimal}}(\{D_i|\forall\,D_i\in\mathcal{D}\})\}$ where $f_{\text{minimal}}$ performs pruning and union operations on multiple sets of definitions to obtain a minimal set. Since any exDefinition can serve as input for a symbolic deduction engine to potentially reach a conclusion, a one-to-many mapping table, called the Knowledge Point-to-exDefinition Mapping Table (K2exD-MT), can be constructed. Therefore, given any knowledge point, the exDefinitions can be obtained through a sampling function: $exD_i = f_{\text{sample}}(K_i, \text{K2exD-MT})$.


\paragraph{\textsc{Definition 4}: Deduced Conclusion.} 
Given several knowledge points and a set of sampled exDefinitions $exD$, different conclusions can be derived by an SDE through step-by-step reasoning. It is not guaranteed that a valid conclusion will always be reached, meaning that some combinations of knowledge points may not lead to a valid conclusion. We treat the \textit{deduced conclusions} $DC$ as the questions of the generated problem in formal language, which are obtained through two functions: $exd = f_{\text{minimal}}(exD)$ and $DC = f_{\text{engine}}(exd)$.


\paragraph{\textsc{Definition 5}: Generated Textual Problem and Diagram.} 
Given a set of exDefinitions $exd$, if a set of deduced conclusions $DC$ is obtained through an SDE, the generated problems in natural language and their corresponding diagram can be derived using two translation functions: $GP^{(\text{text})}_i = f_{\text{text}}(exd, DC_i) = \{CL_i, Q_i\}$ and $GP^{(\text{diagram})} = f_{\text{diagram}}(exd)$, where $CL_i$ and $Q_i$ represent the clauses and the question of the $i$th generated textual problem, respectively.

\paragraph{\textsc{Definition 6}: Geometry Problem Generation Task.} 
Based on the above-mentioned Definitions 1-5, the task of geometry problem generation in this paper is formally defined as follows:
\begin{equation}
\footnotesize
    GP^{(\text{text})},GP^{(\text{diagram})}=f(K,h,\text{K2exD-MT},\text{SDE}),
\end{equation}
where $K$ is the set of knowledge points, $h$ is the difficulty degree, K2exD-MT is the predefined knowledge point-to-exDefinition mapping table, and SDE refers to a symbolic deduction engine.

\begin{figure*}
\centerline{\includegraphics[width=1\textwidth]{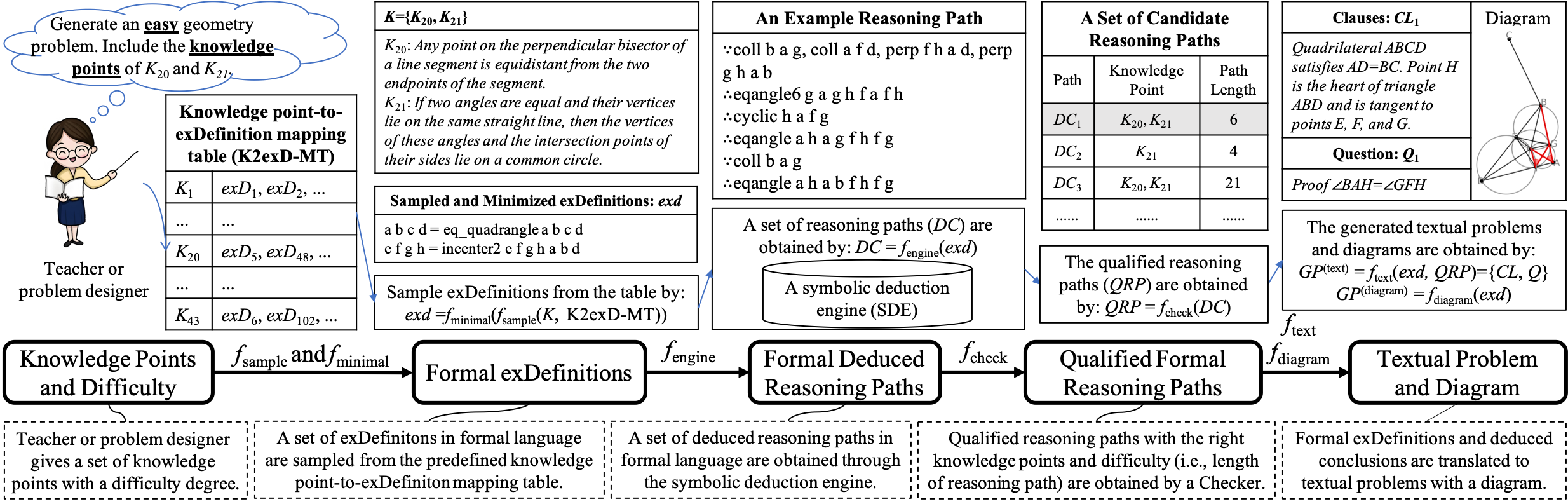}}
\caption{Pipeline of proposed Symbolic Deduction Engine-based Geometry Problem Generation Framework (SDE-PGP) with an example case.} \label{pipeline}
\end{figure*}

\section{Method}\label{method}

In this section, we introduce the pipeline of proposed Symbolic Deduction Engine-based Geometry Problem Generation Framework (SDE-GPG), as shown in Figure~\ref{pipeline}.

\subsection{Offline Construction of Knowledge Point-to-exDefinition Mapping Table}
\label{MappingTable}

As shown in Figure~\ref{pipeline}, our framework relies on a Knowledge Point-to-exDefinition Mapping Table (K2exD-MT), which establishes the relationships between each knowledge point and multiple sets of formal exDefinitions. This way can help to avoid inherent biases in translation between natural and formal languages, which is often faced in solving geometry problems. Algorithm~\ref{alg:algorithm_1} (see Appendix) outlines the process for constructing the table.

In Algorithm 1, two repositories—definitions $\mathcal{D}$\footnote{\url{https://github.com/google-deepmind/alphageometry/blob/main/defs.txt}} and knowledge points $\mathcal{K}$\footnote{\url{https://github.com/google-deepmind/alphageometry/blob/main/rules.txt}}—are leveraged, where $N_d=68$ and $N_k=43$ are their quantities respectively. Given a symbolic deduction engine (SDE) and iteration times $T$, in each iteration, we first sample $n$ definitions from $\mathcal{D}$ to obtain a new set $\hat{D}$. After performing pruning and union operations ($f_{\text{minimal}}$) on $\hat{D}$, a minimal set of definitions, $\hat{d}$, is obtained. Then, the reasoning function ($f_{\text{engine}}$) based on the SDE is executed to generate a set of conclusions $DC$. All knowledge points ${K_i}$ used in the reasoning process are recorded, and a new mapping entry between $K_i$ and $\hat{d}$ is added to the K2exD-MT iteratively. In our primary experiment, we set $n=2$ and $T=100,000$, and the distribution numbers of obtained exDefinition sets corresponding to each knowledge point are shown in Table~\ref{table:K2exD-MT} of the Appendix.



\subsection{K2exD-MT Lookup, exDefinitions Sampling and Symbolic Deduction}
Since the K2exD-MT has been constructed beforehand, during online process, the exDefinitions can be efficiently looked up on the table for each knowledge point. Then, the retrieved exDefinitions can be used to initiate the deduction. In contrast, randomly collecting input definitions from the original repository $\mathcal{D}$ would be inefficient, as the they may be completely unrelated to the given knowledge points. As a result, this method can ensure the proper correlation of the to-be-generated problems with each given knowledge point.

Lines 2-5 of Algorithm 2 (see Appendix) show the process of exDefinitions sampling by using K2exD-MT, while Line 7 represents the deduction process with an SDE. After obtaining the exDefinitions, the $f_{\text{minimal}}$ operation is also performed (Line 6 of Algorithm 2) to obtain a minimal set of exDefinitions before deduction begins. For deduction, we leverage the symbolic engine proposed by AlphaGeometry, retaining all core components of deductive database, algebraic rules, traceback algorithms, and proof pruning~\cite{trinh2024solving}.

\subsection{Problem Qualification Checking}

Although the AlphaGeometry SDE supports the proof pruning, our task is to generate controllable and qualified problems, instead of just data augmentation without caring for the problem's quality. Therefore, an additional function for qualification checking should be developed. After obtaining candidate problems, based on control variables, unqualified problems would be filtered out, which means that the qualified reasoning paths should (1) be shortest paths, (2) involve all the required knowledge points (i.e., completeness of knowledge points), (3) involve all the exDefinitions to reach conclusions (i.e., completeness of clauses), and (4) be consistent with the given difficulty degree (i.e., consistency of difficulty) in terms of the length of paths.
The checking function\footnote{This is an engineering implementation to filter out qualified problems which meet the above four constraints.} is important to ensure the quality of generated problems by filtering out those reasoning paths that are not shortest or incomplete on required control variables.


\subsection{Textual Problem and Diagram Generation}

After obtaining qualified reasoning paths from the previous step, our framework can translate the formal exDefinitions and conclusions into textual problems and diagrams using functions $f_{\text{text}}$ and $f_{\text{diagram}}$, respectively. Lines 8-14 in Algorithm 2 (see Appendix) describe the translation process.

For the translation of textual part, we use a series of predefined templates that can map formal expressions to their corresponding natural language representations, as the grammar of formal language is finite\footnote{All the templates can be published in a code repository.}. An example is shown in Figure~\ref{pipeline}. While the variety of language expressions can be further refined by any LLM, we leave it as a future work.

For the generation of diagrams, due to the specificity of geometry, we implement $f_{\text{diagram}}$ as an iterative process that successively maps each exDefinition $\hat{exd}$ to a geometric diagram using a drawing tool\footnote{\url{https://github.com/google-deepmind/alphageometry/blob/main/graph.py}}. 
These operations are executed sequentially to ensure geometric consistency with the given exDefinitions. For example, point constructions must precede line drawings, and angle markings can only be added once the relevant lines are drawn. The process continues until all geometric statements in $\hat{exd}$ are properly represented in the diagram. Admittedly, sometimes the generated diagrams do not totally align with human conventions, e.g., improper position of a point. A visual interface can be developed to support manual adjustment for users.

\section{Experiment}

In this section, we present the experimental results of our proposed method. Since there are few existing counterparts to serve as baselines and no ground truth available for evaluation, we perform human evaluations focusing on the aspects of readability, solvability and controllability.



\subsection{Dataset}

To address the above questions, we first prepare datasets where each sample should consist of real-world combinations of knowledge points. We curate two datasets of geometry problems in different languages manually. As known, random combinations of knowledge points may not deduce a conclusion. 
In real-world applications, problem designers are typically experts who are familiar with how to meaningfully combine the knowledge points.

\begin{itemize}
\setlength{\itemsep}{0pt}
  \setlength{\parskip}{0pt}
  \setlength{\itemindent}{0em}
\item \textbf{JGEX-AG-231}\footnote{\url{https://www.scribd.com/document/742181523/jgex-ag-231}}: 
The dataset consists of 231 plane geometry problems, offering a diverse range that includes textbook exercises, regional olympiads, and famous geometry theorems. 
Each problem in the dataset is associated with a set of knowledge points, with an average of 9.19 points per problem. For our experiment, we randomly sample fewer than five knowledge points from each problem to reduce complexity.
\item \textbf{GeoQA}\footnote{\url{https://github.com/chen-judge/GeoQA}}: 
The dataset is sourced from authentic middle school exams in China, containing 5,010 geometric problems with detailed annotated solution programs. For our experiment, we randomly select 100 problems from the plane geometry subset, as the SDE we use supports only this topic. We annotate the knowledge points for each problem, with an average of 1.45 knowledge points per problem, indicating that the overall problem's complexity is lower than that in JGEX-AG-231.
\end{itemize}

\begin{table*}
\centering
\footnotesize
\resizebox{\linewidth}{!}{
\begin{tabular}{|l|c|c|c|c|c|c|c|c|}
\hline
\multirow{2}{*}{\makecell[c]{Method}} & \multicolumn{3}{c|}{Readability} & \multicolumn{3}{c|}{Solvability} & \multicolumn{2}{c|}{Controllability} \\
\cline{2-9}
& GF (1-5) & LC (1-5) & DC (1-5) & NS (0-1) & CS (1-5) & CC (0-1) & CKP (0-1) & CD (0-1) \\
\hline
GPT-4o              & 3.05 & 3.60 &   -  & 0.51 & 2.31 & 0.32 & 0.45 & 0.39 \\
\hline
SDE-PGP w/o checking & 3.44 & 3.61 & \textbf{2.61} & 0.72 & 2.51 & 0.53 & 0.53 & 0.40 \\
\hline
SDE-PGP w/ checking  & \textbf{4.25} & \textbf{4.65} & 2.55 & \textbf{1.00} & \textbf{3.55} & \textbf{1.00} & \textbf{0.62} & \textbf{0.63} \\
\hline
\end{tabular}
}
\caption{Average scores for evaluating readability and solvability on JGEX-AG-231 dataset.}\label{JGEX-AG-231}
\end{table*}

\begin{table*}
\centering
\footnotesize
\resizebox{\linewidth}{!}{
\begin{tabular}{|l|c|c|c|c|c|c|c|c|}
\hline
\multirow{2}{*}{\makecell[c]{Method}} & \multicolumn{3}{c|}{Readability} & \multicolumn{3}{c|}{Solvability} & \multicolumn{2}{c|}{Controllability} \\
\cline{2-9}
& GF (1-5) & LC (1-5) & DC (1-5) & NS (0-1) & CS (1-5) & CC (0-1) & CKP (0-1) & CD (0-1) \\
\hline
GPT-4o                & 4.31  & 4.15  &   -  & 0.90 & 3.71 & 0.61 & 0.75 & 0.29 \\
\hline
SDE-PGP w/o checking   & 4.18 & 4.43  & 2.75 & 0.89 & 3.50 & 0.75 & 0.82  & 0.36 \\
\hline
SDE-PGP w/ checking    & \textbf{4.53} & \textbf{4.54}  & \textbf{3.50}  & \textbf{0.96} & \textbf{3.96} & \textbf{0.82} & \textbf{0.94}  & \textbf{0.47} \\
\hline
\end{tabular}
}
\caption{Average scores for evaluating readability and solvability on GeoQA dataset.}\label{GeoQA}
\end{table*}

\subsection{Experimental Design}


\subsubsection{Measurement Metrics}
\paragraph{Readability.} The generated geometry problems should be humanly-readable, and the evaluation dimensions are as follows:
\begin{itemize}
\setlength{\itemsep}{0pt}
  \setlength{\parskip}{0pt}
  \setlength{\itemindent}{0em}
    \item Grammatical Fluency (GF): It assesses how grammatically clear and concise the language is, and whether there are any ambiguous or confusing expressions.
    \item Logical Correctness (LC): It evaluates the logical structure of the problem, ensuring information is presented in a coherent and orderly manner (e.g., a point should be introduced only after the corresponding line is drawn).
    \item Diagram Correctness (DC): It examines the logical consistency between the textual description and the diagram, and whether the diagram is easily interpretable by humans.
\end{itemize}

\paragraph{Solvability.} The generated geometry problems and diagrams should be solvable, and all the relevant clauses should be incorporated. The evaluation dimensions include:
\begin{itemize}
\setlength{\itemsep}{0pt}
  \setlength{\parskip}{0pt}
  \setlength{\itemindent}{0em}
    \item Native Solvability (NS): Whether the generated problem can be solved.
    \item Consistent Solvability (CS): How well the textual content, the reference answer, and the diagram align to solve the problem, and whether the reasoning path is shortest.
    \item Completeness of Clauses (CC): Whether all clauses are utilized in solving the problem.
\end{itemize}

\paragraph{Controllability.} The generated problems should support that all the required control variables, i.e., knowledge points and difficulty degree in this paper, are satisfied. The dimensions include:

\begin{itemize}
\setlength{\itemsep}{0pt}
  \setlength{\parskip}{0pt}
  \setlength{\itemindent}{0em}
    \item Completeness of Knowledge Points (CKP): Whether all the required knowledge points are involved in solving the problem.
    \item Consistency of Difficulty (CD): Whether the length of reasoning path is consistent with the required difficulty degree. We empirically set Easy for less than 10 steps, Moderate for between 10 and 20 steps, and Difficult for larger than 20 steps.
\end{itemize}



\subsubsection{Measurement Method}

For evaluating the metrics of readability, solvability and controllability, human annotation is conducted. We invite three experts with substantial experience in geometry problem design, two of whom serve as the initial judges and another one as the arbiter. When the results from the judges are inconsistent, the arbiter makes the final decision. We use two types of scoring: a discrete grading score ranging from 1 to 5 (orderly corresponding to poor, wrong, fair, good, perfect), and a binary score of 0 or 1 (0 is negative and 1 is positive). The grading score is used to measure GF, LC, DC, and CS, while the binary score is for NS, CC, CKP and CD. We report the average scores for both datasets, respectively.

We use GPT-4o\footnote{\url{https://chatgpt.com/}} and SDE-PGP without checking as baselines, and write a prompt for the LLM to generate geometry problems (see Table~\ref{prompt_geo} in Appendix). Note that current LLMs mostly cannot draw geometric diagrams. For each given input test sample, we generate only one problem and use it for evaluation, rather than generating multiple times to select the best one.

\subsection{Results and Analysis}

\paragraph{Results for Readability.} 
From Table~\ref{JGEX-AG-231} and Table~\ref{GeoQA}, we can see that the generated problems remain generally readable across both datasets. In particular, SDE-PGP w/ checking achieves the highest GF (General Fluency) and LC (Linguistic Clarity) on both datasets, indicating that introducing the checking function leads to more coherent and fluent texts. The DC scores may suggest that SDE-PGP w/o checking may generate easier problems, leading to drawing better diagrams.

\paragraph{Results for Solvability.} 
From Table~\ref{JGEX-AG-231} and Table~\ref{GeoQA}, several observations can be made regarding the metric of solvability: (1) SDE-PGP w/ checking achieves near-perfect Native Solvability (NS), with 1.00 on JGEX-AG-231 and 0.96 on GeoQA, indicating that almost all generated problems are solvable. (2) The Consistent Solvability (CS) score tends to be higher on GeoQA, possibly because the reduced number of knowledge points makes diagram construction and text–diagram consistency easier. (3) The completeness of clauses (CC) is sufficiently high for SDE-PGP w/ checking (1.00 on JGEX-AG-231 and 0.82 on GeoQA), though there remains room for enhancing clause generation in future improvement.

\paragraph{Results for Controllability.}
From Table~\ref{JGEX-AG-231} and Table~\ref{GeoQA}, SDE-PGP w/ checking consistently achieves higher completeness of knowledge points (CKP) and consistency of difficulty (CD) than the baselines on both datasets, validating the effectiveness of the proposed checking function.


\subsection{Case Study}
We provide several representative examples to illustrate the strengths and limitations of our SDE-GPG framework. These examples highlight the framework's effectiveness in generating geometry problems that are readable, solvable, and controllable, as well as identifying areas where further improvement is needed. For detailed discussions and visual examples, please refer to Appendix~\ref{case_study}.

\section{Conclusion}

In this paper, we introduce a novel task of generating readable and solvable geometry problems under the constraint of control variables. To achieve this, we leverage a symbolic deduction engine and propose a new framework called the Symbolic Deduction Engine-based Geometry Problem Generation Framework (SDE-GPG). By creating a mapping table between knowledge points and definitions, our framework eliminates inherent biases in translating natural language into formal language. Our method highlights a checking function to guarantee the problem quality and controllability, as well as enabling the generation of multi-modal geometry problems. The thorough experiments demonstrate the effectiveness of our method on all the readability, solvability and controllability. In the future, situations that involve more control variables, such as context and problem type, and geometric topics, such as geometric inequalities and combinatorial geometry, could be further explored.

\section*{Acknowledgment}
This work is supported by the Special Program on Education Examinations of the China Education Development Strategy Society (Grant No. jyks2024038), the Program of Shanghai Committee of Science and Technology, China (Grant No. 24511103200), and the International Science and Technology Cooperation Program of Shanghai Committee of Science and Technology, China (Grant No. 24170790602). 
We thank all the anonymous reviewers for their insightful and constructive comments. 
Zhuoxuan Jiang is the corresponding author. 

\bibliography{ijcai25}

\clearpage
\appendix
\onecolumn
\section*{Appendix}\label{appendix}

\section{Case Study}
\label{case_study}
As shown in Example 1, it demonstrates a geometry problem generated with our complete SDE-GPG framework, incorporating the checking function. From the perspective of readability, the textual description is clear, grammatically fluent, and logically coherent. The clauses introduce each geometric element sequentially, ensuring logical correctness and clarity. Regarding solvability, the reasoning path is explicit, shortest, and fully utilizes all clauses.

As presented in Example 2, it is generated without using our checking function. Although this problem still maintains decent readability and solvability, the textual description remains fluent, and the diagram clearly corresponds to the textual information, it notably lacks in controllability. Specifically, the generated problem is overly simplified, resulting in a very short reasoning path. Consequently, the actual difficulty is significantly lower than the predefined control variable. This highlights the essential role of our checking function in controlling and ensuring the complexity and completeness of generated geometry problems.

As shown in Example 3, it represents one of the occasional problematic outputs of our method. Despite having high readability in terms of grammar and logical structure, the generated problem suffers significantly from solvability issues. The main reason for this issue is the absence of certain intermediate theorems within the symbolic deduction engine. As a result, the system performs unnecessarily lengthy deductions for a conclusion that could ideally be derived in just a single step. This leads to a non-shortest reasoning path. To address this issue in future work, we plan to enrich our symbolic deduction engine with additional intermediate geometric theorems, further optimizing the efficiency of our geometry problem generation framework.

Example 4 illustrates an incorrect geometry problem generated by GPT-4o. This example highlights typical errors encountered when relying solely on LLMs for geometry problem generation, such as logical errors in the problem formulation, incorrect or impossible-to-solve scenarios, and the improper application of geometric theorems. Such issues underscore the importance of integrating symbolic deduction engines and rigorous checking mechanisms, as proposed by our SDE-GPG framework.

\onecolumn
\begin{tcolorbox}[colback=gray!10!white,colframe=black,title=Example 1: An ideal geometry problem generated by SDE-GPG with checking.]
\textbf{Problem:}
Let points \( A, B \) define segment \( AB \). Let point \( C \) be the midpoint of segment \( BA \). Construct point \( D \) as the reflection of \( C \) about point \( B \). Let point \( E \) lie on both the circle centered at \( C \) with radius \( CA \), and the circle centered at \( B \) with radius \( BC \). Construct point \( F \) such that \( BF \perp AB \) and point \( F \) lies on line \( AE \). Construct point \( G \) such that \( G \) lies on both line \( BF \) and line \( DE \).

The following conditions hold:  
\begin{itemize}
    \item Points \( B, C, A \) are collinear, and \( CB = CA \).
    \item Points \( B, C, D \) are collinear, and \( BC = BD \).
    \item \( CE = CA \), \( BE = BC \).
    \item Points \( E, F, A \) are collinear.
    \item \( BF \perp AB \).
    \item Points \( E, G, D \) are collinear, and points \( F, B, G \) are collinear.
\end{itemize}

Prove: 
The angle formed between lines \( AE \) and \( BF \) equals the angle formed between lines \( DE \) and \( CG \).

\begin{center}
    \includegraphics[width=0.45\textwidth]{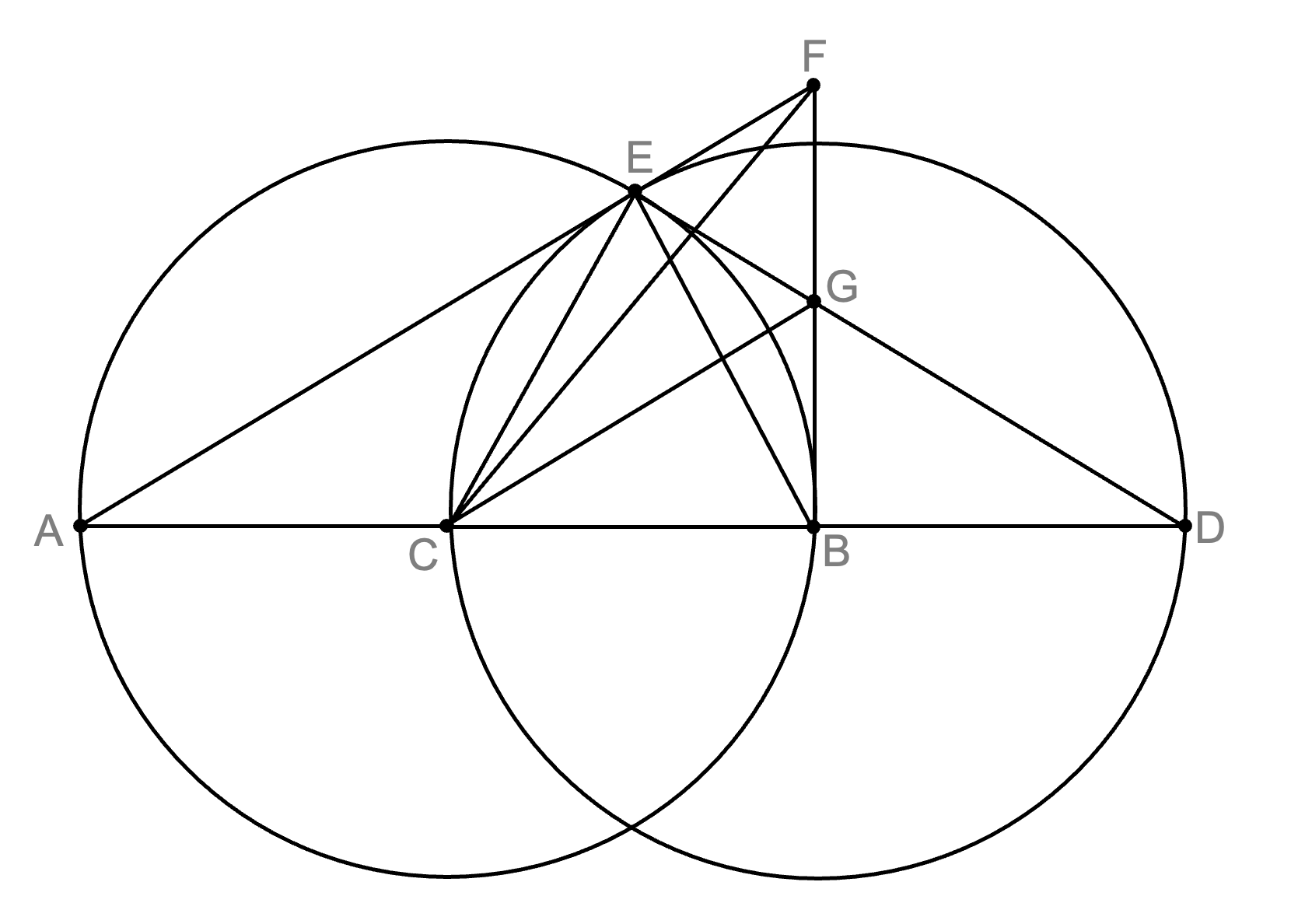}
\end{center}

\hrulefill

{\small
\textbf{Proof Steps:}
\[
\begin{aligned}
&(1)\quad CE = CA,\; CB = CA \implies C \text{ is the circumcenter of } \triangle BEA. \\
&(2)\quad C \text{ is circumcenter of } \triangle BEA,\; B,C,A \text{ collinear} \implies BE \perp AE. \\
&(3)\quad BC = BD,\; \angle DBG = \angle GBC \implies \angle BDG = \angle GCB. \\
&(4)\quad BC = BD,\; BE = BC \implies BE = BD. \\
&(5)\quad BE = BD \implies \angle BED = \angle EDB. \\
&(6)\quad G,D,E \text{ collinear},\; B,C,D \text{ collinear},\; B,C,A \text{ collinear},\; \angle BDG = \angle GCB,\; \angle BED = \angle EDB \\ &\quad \implies \angle BEG = \angle(\text{line }BD,\text{line }GC). \\
&(7)\quad \angle FEB = \angle FBD,\; \angle BEG = \angle(\text{line }BD,\text{line }GC) \implies \angle FEG = \angle(\text{line }FB,\text{line }GC). \\
&(8)\quad \angle FEG = \angle(\text{line }FB,\text{line }GC),\; E,F,A\text{ collinear},\; E,G,D\text{ collinear} \\& \quad \implies\angle(AE,BF)=\angle(DE,CG).
\end{aligned}
\]
}

Thus, the proof is completed:
\[
\angle(AE,BF)=\angle(DE,CG)
\]
\end{tcolorbox}

\begin{tcolorbox}[colback=gray!10!white,colframe=black,title=Example 2: A geometry problem generated by SDE-GPG without checking.]
\textbf{Problem:}
Construct a triangle \(\triangle ABC\). Let points \(D, E, F\) be the midpoints of segments \(CB, AB, AC\), respectively. Point \(G\) is positioned such that distances from \(G\) to points \(D, E, F\) are all equal.  
Prove that the angle formed by line \(DG\) and side \(AB\) is equal to the angle formed by side \(AB\) and line \(FG\).

\begin{center}
    \includegraphics[width=0.4\textwidth]{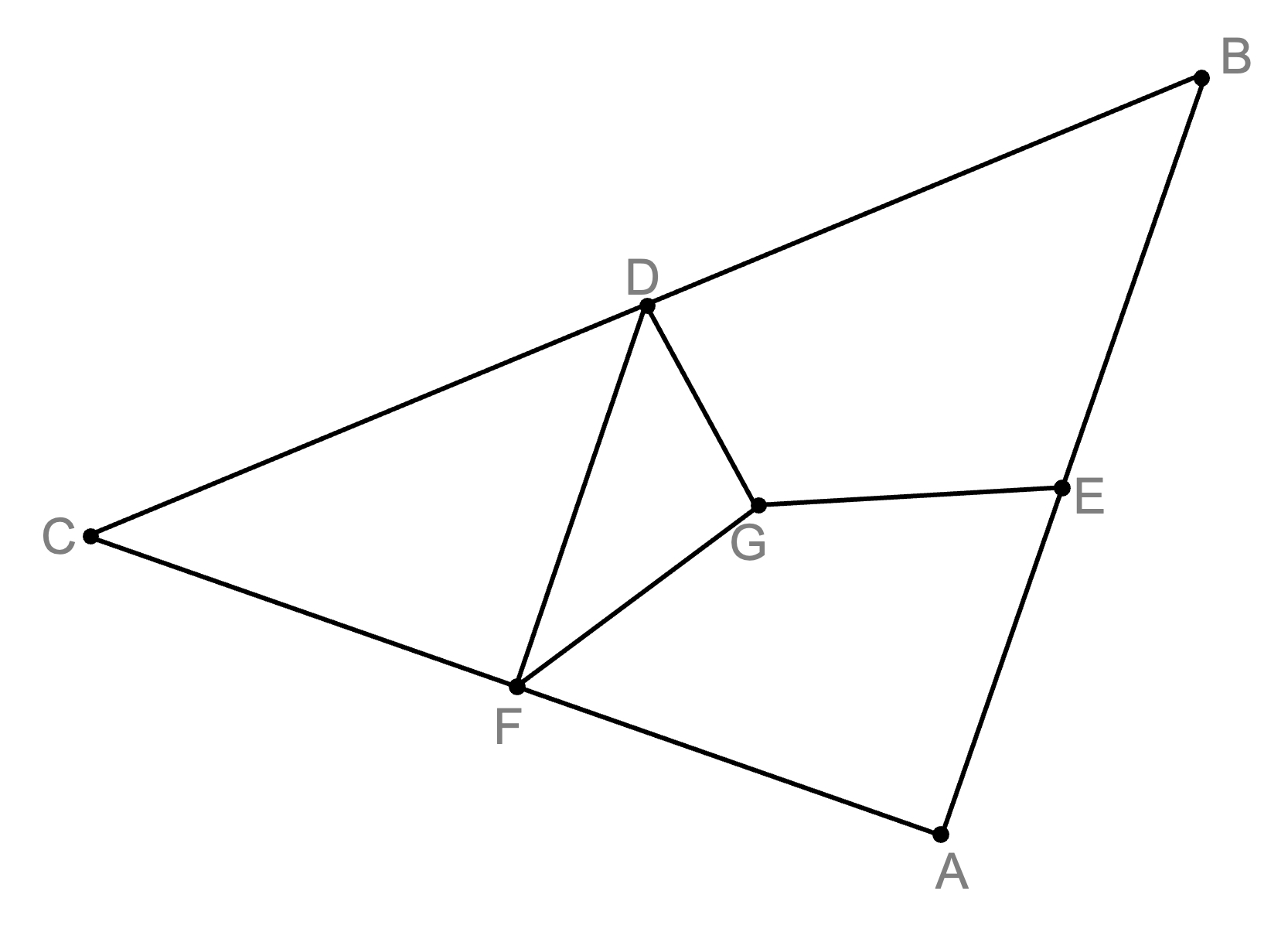}
\end{center}

\hrulefill

\textbf{Proof Steps:}
\[
\begin{aligned}
&(1)\quad GD = GF \implies \angle GDF = \angle DFG. \\
&(2)\quad F \text{ is the midpoint of } AC,\; D \text{ is the midpoint of } BC \implies FD \parallel AB. \\
&(3)\quad \angle GDF = \angle DFG,\; FD \parallel AB \implies \angle(DG, AB) = \angle(AB, FG).
\end{aligned}
\]

Thus, the proof is completed:
\[
\angle(DG, AB) = \angle(AB, FG)
\]

\end{tcolorbox}

\begin{tcolorbox}[colback=gray!10!white,colframe=black,title=Example 3: A problematic geometry problem generated due to missing intermediate theorems.]
\textbf{Problem:}

Construct a square \(ABCD\). Let point \(E\) be the intersection point of diagonals \(CA\) and \(BD\).  
Prove:
\[
\frac{BE}{BD} = \frac{CE}{BD}
\]

\begin{center}
    \includegraphics[width=0.45\textwidth]{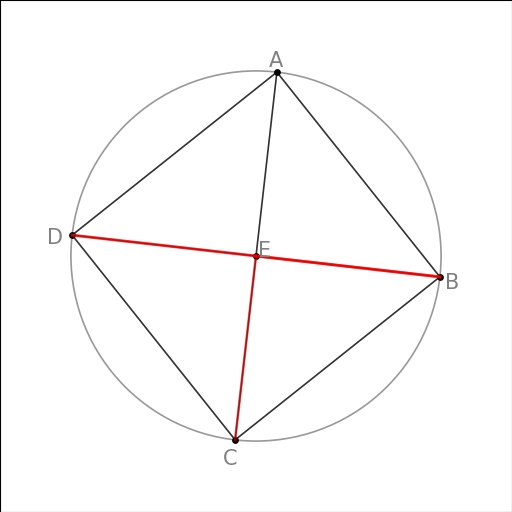}
\end{center}

\hrulefill

\textbf{Proof Steps:}

\[
\begin{aligned}
&(1)\quad AD \parallel BC,\; B,D,E \text{ are collinear},\; C,A,E \text{ are collinear} \implies \frac{BE}{BD} = \frac{CE}{CA}.\\
&(2)\quad AB \perp BC,\; AB \parallel CD \implies BC \perp CD.\\
&(3)\quad AB \perp BC,\; BC \parallel AD \implies DA \perp AB.\\
&(4)\quad AB \perp BC,\; BC \perp CD,\; DA \perp AB \implies AC = BD.\\
&(5)\quad \frac{BE}{BD} = \frac{CE}{CA},\; AC = BD \implies \frac{BE}{BD} = \frac{CE}{BD}.
\end{aligned}
\]

Thus, the proof is completed:
\[
\frac{BE}{BD} = \frac{CE}{BD}
\]
\end{tcolorbox}

\begin{tcolorbox}[colback=gray!10!white,colframe=black,title=Example 4: A geometry problem generated by GPT-4o demonstrating inaccuracies.]
\textbf{Problem:}

Given triangles \(\triangle ABC\) and \(\triangle DEF\), with \(AB = DE\), \(AC = DF\), and \(\angle BAC = \angle EDF\):

(1) Prove that \(\triangle ABC \cong \triangle DEF\).

(2) If \(BC = 8\), \(AB = 6\), and \(\angle EDF = 60^\circ\), find the perimeter of \(\triangle DEF\). 

\hrulefill

\textbf{Proof Steps:}

(1) Proving \(\triangle ABC \cong \triangle DEF\):

According to the given conditions:

\begin{itemize}
\item \( AB = DE \) (corresponding sides are equal),
\item \( AC = DF \) (corresponding sides are equal),
\item \( \angle BAC = \angle EDF \) (included angles are equal).
\end{itemize}

By the congruence criterion (\textbf{SAS Criterion}), triangles having two corresponding sides and their included angles equal are congruent. Thus, we have:
\[
\triangle ABC \cong \triangle DEF.
\]

(2) Finding the perimeter of \(\triangle DEF\):

Since \(\triangle ABC \cong \triangle DEF\), the corresponding sides of these two triangles are equal, thus \( BC = EF \).

Given:
\[
BC = 8,\quad AB = 6,\quad DE = AB = 6.
\]

Also, by congruence properties:
\[
AC = DF = 6.
\]

Therefore, the side lengths of \(\triangle DEF\) are:
\[
DE = 6,\quad DF = 6,\quad EF = 8.
\]

The perimeter of \(\triangle DEF\) is:
\[
DE + DF + EF = 6 + 6 + 8 = 20.
\]

Answer:

(1) \(\triangle ABC \cong \triangle DEF\);

(2) The perimeter of \(\triangle DEF\) is \(20\).

\end{tcolorbox}

\twocolumn

\onecolumn
\section{Detailed Information about SDE-GPG}

\begin{algorithm}
	\caption{Constructing the knowledge point-to-exDefinition mapping table (K2exD-MT)}
	\label{alg:algorithm_1}
	\KwInput{The repository of definitions $\mathcal{D}$, the repository of knowledge points $\mathcal{K}$, the SDE, the iteration times $T$}
	\KwOutput{K2exD-MT}
        K2exD-MT=$\{\}$, t=1\;
	\While{$t<T$}{
            Sample an integer $n\in\{1,N_d\}$ and sample $n$ definitions from $\mathcal{D}$ to construct a new set $\hat{D}$\;
            $\hat{d}=f_{\text{minimal}}(\hat{D})$\;
            $DC=f_{\text{engine}}(\hat{d})$\;
            Record all the knowledge points $\{K_i\}$ used along with the reasoning paths from $\hat{d}$ to any $DC_i$\;
            \ForEach{$K_i\in\{K_i\}$}{
                Insert one mapping of [$K_i\to \hat{d}$] into K2exD-MT\;
            }
            t=t+1\;
        }
	\textbf{return} K2exD-MT.
\end{algorithm}

\begin{algorithm}
	\caption{Generating geometry problems}
	\label{alg:algorithm_2}
	\KwInput{A set of knowledge points $\hat{K}$, a difficulty degree $h$, the K2exD-MT, the SDE}
	\KwOutput{$GP^{(\text{text})}$, $GP^{(\text{diagram})}$}
	$GP^{(\text{text})}=\{\}$, $GP^{(\text{diagram})}=\{\}$, $\hat{exD}=\{\}$\;
        \ForEach{$K_i\in\hat{K}$}{
            $exD_i=f_{\text{sample}}(K_i,\text{K2exD-MT})$\;
            $\hat{exD}=\hat{exD}+\{exD_i\}$\;
        }
        $\hat{exd}=f_{\text{minimal}}(\hat{exD})$\;
        $\hat{QRP}=f_{\text{check}}(f_{\text{engine}}(\hat{exd}))$\;
        \If{$\hat{QRP}\neq\{\}$}{
            $GP^{(\text{diagram})}=\{f_{\text{diagram}}\{\hat{exd}\}$\}\;
            \ForEach{$QRP_i\in\hat{QRP}$}{
                $GP^{(\text{\text{text}})}_i=f_{\text{text}}\{\hat{exd},QRP_i\}$\;
                $GP^{(\text{text})}=GP^{(\text{text})}+\{GP^{(\text{text})}_i\}$\;
            }
        }
	\textbf{return} $GP^{(\text{text})}$ and $GP^{(\text{diagram})}$.
\end{algorithm}

\begin{table*}[hb]
    \centering
    \begin{tabular}{|p{0.2\textwidth}|c|p{0.58\textwidth}|}
        \hline
         \textbf{Term} & \textbf{Notation} & \textbf{Description} \\
        \hline
        Clauses & $CL$ & The clauses of a textual problem. \\
        \hline
        Question & $Q$ & The question of a textual problem. \\
        \hline
        Textual Problem & $\{CL,Q\}$ & A paragraph of problem description including clauses and a question. \\
        \hline
        Diagram  & - & A corresponding geometric diagram for a textual problem. \\
        \hline
        Knowledge points & $\mathcal{K}$ & A control variable that corresponds to geometric rules, including theorems and properties. The scope is finite. \\
        \hline
        The number of knowledge points & $N_k$ & The number of knowledge points in an existing repository.\\
        \hline
        Difficulty Degree & $h$ & A control variable where its scope is empirically set as Easy for less than 10 reasoning steps, Moderate for 10 to 20 steps, and Difficulty for larger than 20 steps. \\
        \hline
        Premises & $P$ & The part of clauses of a knowledge point in formal language. \\
        \hline
        Conclusion & $C$ & The part of conclusion of a knowledge point in formal language.  \\
        \hline
        Definitions & $\mathcal{D}$ & A set of complete formal descriptions of geometry to start deduction on a symbolic deduction engine.\\
        \hline
        The number of definitions & $N_d$ & The number of definitions in an existing repository.\\
        \hline
        Extended Definitions (exDefinitions) & $ex\mathcal{D}$ & A repository including all the combination of any definitions.  \\
        \hline
        Knowledge Point-to-exDefinition Mapping Table & K2exD-MT & A mapping table between knowledge points to exDefinitions. \\
        \hline
        Deduced Conclusion & $DC$ & A conclusion deduced by using a symbolic deduction engine given a set of extended definitions. \\
        \hline
        Qualified Reasoning Path & $QRP$ & Qualified reasoning paths by using a checking function to ensure the quality and controllability. \\
        \hline
        Symbolic Deduction Engine & SDE & An engine which can automatically deduce by inputting some definitions in specific formal language.\\
        \hline
        Generated Textual Problem & $GP^{(\text{text})}$ & A set of textual problems generated by SDE-GPG. \\
        \hline
        Generated Diagram & $GP^{(\text{diagram})}$ &  A geometric diagram generated by SDE-GPG. \\
        \hline
        Sample Function & $f_{\text{sample}}$ & A function to sample a set of exDefinitions from K2exD-MT by given a knowledge point. \\
        \hline
        Minimal Function & $f_{\text{minimal}}$ & A function to perform pruning and union operations on multiple sets of definitions or exDefinitions to obtain a minimal set. \\
        \hline
        Engine Function & $f_{\text{engine}}$ & A function to deduce reasoning paths from given definitions or exDefinitions to a set of deduced conclusions, including core components of Deductive Database (DD), Algebraic Rules (AR), traceback algorithms, and proof pruning. \\
        \hline
        Checking Function & $f_{\text{check}}$ & A function to filter out unqualified reasoning paths based on given control variables. \\
        \hline
        Text Function & $f_{\text{text}}$ & A function to translate exDefinitions and deduced conclusions from formal language to natural language. \\
        \hline
        Diagram Function & $f_{\text{diagram}}$ & A function to translate geometric  exDefinitions to a diagram. \\
        \hline
    \end{tabular}
    \caption{Description of terms and notations used in this paper.}\label{table:term}
\end{table*}

\setlength{\LTcapwidth}{\textwidth}
\onecolumn

\begin{longtable}{|c|l|p{0.35\textwidth}|p{0.11\textwidth}|}
\hline
\textbf{ID} & \textbf{Knowledge Point Code} & \textbf{Description} & \textbf{No. of exDefinition Sets} \\
\hline

        $K_1$  & eqangle6\_eqangle6\_ncoll\_cong\_contri2 & If two triangles have two angles and the corresponding non-included side equal, then the two triangles are congruent. & 10,435 \\
        \hline
        $K_2$  & eqratio6\_eqratio6\_ncoll\_simtri* & If two triangles have their corresponding sides in proportion and the included angle equal, then the two triangles are similar. & 13,232 \\
        \hline
        $K_3$  & cong\_cong\_eqangle6\_ncoll\_contri* & If two triangles have two sides and the included angle equal, then the two triangles are congruent. & 12,108 \\
        \hline
        $K_4$  & eqratio6\_eqratio6\_ncoll\_cong\_contri* & If the segments $BA:BC = QP:QR$ and $CA:CB = RP:RQ$, and points $A$, $B$, and $C$ are not collinear, and $AB = PQ$, then $\angle ABC$ and $\angle PQR$ are congruent. & 12,108 \\
        \hline
        $K_5$  & eqratio6\_eqangle6\_ncoll\_simtri* & If two triangles have their corresponding sides in proportion and the included angle equal, then the two triangles are similar. & 13,232 \\
        \hline
        $K_6$  & eqangle6\_eqangle6\_ncoll\_simtri2 & If two triangles have their corresponding angles equal, then the two triangles are similar. & 10,948 \\
        \hline
        $K_7$  & eqangle6\_ncoll\_cong & If two angles of a triangle are equal, then the triangle is an isosceles triangle. & 8,681 \\
        \hline
        $K_8$  & cong\_ncoll\_eqangle & In an isosceles triangle, the base angles are equal. & 8,681 \\
        \hline
        $K_9$  & cong\_cong\_cong\_ncoll\_contri* & If two triangles have their corresponding three sides equal, then the two triangles are congruent. & 12,108 \\
        \hline
        $K_{10}$  & eqangle6\_eqangle6\_ncoll\_simtri & If two triangles have their corresponding two angles equal, then the two triangles are similar. & 10,205 \\
        \hline
        $K_{11}$  & eqangle6\_eqangle6\_ncoll\_cong\_contri & If two triangles have their corresponding two angles and the included side equal, then the two triangles are congruent. & 8,613 \\
        \hline
        $K_{12}$  & eqangle\_eqangle\_eqangle & If the angles between two pairs of lines are equal, then the angles between these two pairs of lines are transitive. & 20,644 \\
        \hline
        $K_{13}$  & eqangle\_perp\_perp & If the angle between $AB$ and $PQ$ is equal to the angle between $CD$ and $UV$, and $PQ$ is perpendicular to $UV$, then $AB$ is perpendicular to $CD$. & 26,733 \\
        \hline
        $K_{14}$  & circle\_eqangle\_perp & If $O$ is the circumcenter of triangle $ABC$ and $\angle BAX = \angle BCA$, then $OA$ is perpendicular to $AX$. & 2,705 \\
        \hline
        $K_{15}$  & cong\_cong\_cyclic\_perp & If $AP = BP$, $AQ = BQ$, and quadrilateral $ABPQ$ is cyclic, then $PA$ is perpendicular to $AQ$. & 3,170 \\
        \hline
        $K_{16}$  & cyclic\_eqangle\_cong & In the same circle, if two inscribed angles are equal, then the chords subtended by these angles are equal. & 8,289 \\
        \hline
        $K_{17}$  & perp\_perp\_npara\_eqangle & If two lines are perpendicular to two other lines, and these two lines are not parallel, then the angles between them are equal. & 19,540 \\
        \hline
        $K_{18}$  & cong\_cong\_perp & If a point is equidistant from the two endpoints of a line segment, then the point lies on the perpendicular bisector of the line segment. & 5,372 \\
        \hline
        $K_{19}$  & circle\_perp\_eqangle & If $O$ is the circumcenter of triangle $ABC$ and $OA$ is perpendicular to $AX$, then $\angle BAX = \angle BCA$. & 2,705 \\
        \hline
        $K_{20}$  & cyclic\_eqangle & In the same circle, inscribed angles subtended by the same arc or equal arcs are equal. & 8,289 \\
        \hline
        $K_{21}$  & eqangle6\_ncoll\_cyclic & If two angles are equal and their vertices lie on the same straight line, then the vertices of these angles and the intersection points of their sides lie on a common circle. & 8,289 \\
        \hline
        $K_{22}$  & eqratio\_coll\_coll\_ncoll\_sameside\_para & If $OA:AC = OB:BD$, and $O, A, C$ are collinear, $O, B, D$ are collinear, $A, B, C$ are not collinear, and $A, O, C$ and $B, O, D$ are on the same side, then $AB$ is parallel to $CD$. & 913 \\
        \hline
        $K_{23}$  & para\_coll & If two lines are parallel, they have no common points unless they are the same line. & 7,421 \\
        \hline
        $K_{24}$  & para\_coll\_coll\_eqratio3 & If two parallel lines are intersected by two transversal lines, then the corresponding line segments formed are proportional. & 1,013 \\
        \hline
        $K_{25}$  & midp\_midp\_para\_1 & The midline of a triangle is parallel to the third side. & 570 \\
        \hline
        $K_{26}$  & eqratio\_eqratio\_eqratio & If two proportions are equal and their middle terms are also equal, then other proportional relationships can be proved by the transitivity of proportions. & 2,728 \\
        \hline
        $K_{27}$  & eqangle\_para & If two lines are intersected by a third line and the alternate interior angles are equal, then the two lines are parallel. & 2,682 \\
        \hline
        $K_{28}$  & cyclic\_para\_eqangle & If quadrilateral $ABCD$ is cyclic and $AB$ is parallel to $CD$, then $\angle ADC = \angle BCD$. & 6,216 \\
        \hline
        $K_{29}$  & eqratio6\_coll\_ncoll\_eqangle6 & If the ratio of the distances from a point to two sides of a triangle is equal to the ratio of those two sides, then the point lies on the angle bisector. & 2,170 \\
        \hline
        $K_{30}$  & eqangle6\_coll\_ncoll\_eqratio6 & If a point lies on the angle bisector of a triangle, then the ratio of its distances to the two sides of the triangle is equal to the ratio of those two sides. & 2,169 \\
        \hline
        $K_{31}$  & circle\_coll\_perp & In a circle, the inscribed angle subtended by the diameter is a right angle. & 1,453 \\
        \hline
        $K_{32}$  & perp\_midp\_cong & In a right-angled triangle, the median to the hypotenuse is half the length of the hypotenuse. & 1,451 \\
        \hline
        $K_{33}$  & eqratio\_cong\_cong & If two proportions are equal, and one pair of corresponding line segments are equal, then the other pair of corresponding line segments are also equal. & 464 \\
        \hline
        $K_{34}$  & para\_coll\_coll\_para\_eqratio6 & If $AB$ is parallel to $CD$, $M, A, D$ are collinear, $N, B, C$ are collinear, and $MN$ is parallel to $AB$, then $MA:MD$ = $NB:NC$. & 233 \\
        \hline
        $K_{35}$  & midp\_midp\_eqratio & If a point is the midpoint of a line segment, then it divides the segment into two equal parts. & 257 \\
        \hline
        $K_{36}$  & midp\_perp\_cong & Any point on the perpendicular bisector of a line segment is equidistant from the two endpoints of the segment. & 1,805 \\
        \hline
        $K_{37}$  & perp\_perp\_ncoll\_para & If two lines are both perpendicular to the same line, then these two lines are parallel. & 278 \\
        \hline
        $K_{38}$  & para\_coll\_coll\_eqratio6\_sameside\_para & If $AB$ is parallel to $CD$, $M, A, D$ are collinear, $N, B, C$ are collinear, $MA:MD = NB:NC$, and $M, A, D$ and $N, B, C$ are on the same side, then MN is parallel to $AB$. & 234 \\
        \hline
        $K_{39}$  & cong\_cong\_cong\_cyclic & If a point is equidistant from the four vertices of a quadrilateral, then the four vertices of the quadrilateral lie on a common circle. & 466 \\
        \hline
        $K_{40}$  & circle\_coll\_eqangle\_midp & If $O$ is the circumcenter of triangle $ABC$, $M, B, C$ are collinear, and $\angle BAC = \angle BOM $, then $M$ is the midpoint of $BC$. & 190 \\
        \hline
        $K_{41}$  & circle\_midp\_eqangle & If $O$ is the circumcenter of triangle $ABC$ and $M$ is the midpoint of $BC$, then $\angle BAC = \angle BOM$. & 192 \\
        \hline
        $K_{42}$  & midp\_midp\_para\_2 & If $M$ is the midpoint of $AB$ and also the midpoint of $CD$, then $AC$ is parallel to $BD$. & 329 \\
        \hline
        $K_{43}$  & midp\_para\_para\_midp & In a parallelogram, the diagonals bisect each other. & 327 \\
        \hline
\caption{%
  Statistics of the knowledge point-to-definition mapping table (K2exD-MT). The knowledge point codes (or rule codes) follow the settings of AlphaGeometry. The detailed table data including the expressions in formal language will be published in a public code repository.
} \label{table:K2exD-MT}
\end{longtable}

\twocolumn


\begin{table*}
\centering
\begin{tabular}{p{\textwidth}}
\hline
Please generate a high-quality question based on the following knowledge point: \\

Knowledge Point: $<$content$>$ \\
Make sure the generated question meets the following requirements: \\

1. Accurately reflects the specified knowledge point and assesses the student's understanding and ability to apply it \\
2. The wording of the question should be clear and unambiguous, conforming to academic standards \\
3. The difficulty level should be moderate, with a certain degree of thinking value and differentiation \\
4. The question should include a clear problem-solving approach and a standard answer \\
The content should be original and avoid using common examples or exercises \\
Please output in the following format: \\
\textbf{Question} \\
(Provide the full description of the question here) \\
\textbf{Explanation} \\
(Provide a detailed solution process and answer explanation here) \\
\hline
\end{tabular}
\caption{Prompt template used for geometry problem generation with LLMs.}\label{prompt_geo}
\end{table*}

\end{document}